\documentclass{article}

     \usepackage[preprint,nonatbib]{neurips_2020}

\usepackage[utf8]{inputenc} 
\usepackage[T1]{fontenc}    
\usepackage{hyperref}       
\usepackage{url}            
\usepackage{booktabs}       
\usepackage{amsfonts}       
\usepackage{nicefrac}       
\usepackage{microtype}      
\usepackage{subfig}
\usepackage{amsmath}
\usepackage{amssymb}
\usepackage{amsthm}
\usepackage{mathtools}
\usepackage{stfloats}
\usepackage{multirow}
\usepackage[toc,page]{appendix}
\usepackage{wrapfig}
\usepackage{microtype}      

\newcommand{\eg}{\emph{e.g. }}
\newcommand{\etal}{\emph{et.al.}}
\newcommand{\ie}{\emph{i.e. }}
\newcommand{\etc}{\emph{etc}}
\newcommand{\wrt}{\emph{w.r.t. }}

\usepackage{hyperref}
\usepackage{algorithm}
\usepackage{algpseudocode}

\title{DC-NAS: Divide-and-Conquer Neural Architecture Search}

\author{Yunhe Wang\thanks{Authors contributed equally to this work.}$^{\ \ 1}$, Yixing Xu$^{*1}$, Dacheng Tao$^2$\\
	$^1$Noah's Ark Lab, Huawei Technoligies\\
	$^2$The University of Sydney, Darlington, NSW 2008, Australia\\
	\texttt{\{yunhe.wang, yixing.xu\}@huawei.com} \\
	\texttt{dacheng.tao@sydney.edu.au}
}

\begin{document}

\maketitle

\begin{abstract}
	Most applications demand high-performance deep neural architectures costing limited resources. Neural architecture searching is a way of automatically exploring optimal deep neural networks in a given huge search space. However, all sub-networks are usually evaluated using the same criterion; that is, early stopping on a small proportion of the training dataset, which is an inaccurate and highly complex approach. In contrast to conventional methods, here we present a divide-and-conquer (DC) approach to effectively and efficiently search deep neural architectures. Given an arbitrary search space, we first extract feature representations of all sub-networks according to changes in parameters or output features of each layer, and then calculate the similarity between two different sampled networks based on the representations. Then, a k-means clustering is conducted to aggregate similar architectures into the same cluster, separately executing sub-network evaluation in each cluster. The best architecture in each cluster is later merged to obtain the optimal neural architecture. Experimental results conducted on several benchmarks illustrate that DC-NAS can overcome the inaccurate evaluation problem, achieving a $75.1\%$ top-1 accuracy on the ImageNet dataset, which is higher than that of state-of-the-art methods using the same search space.
\end{abstract}

\section{Introduction}
Deep neural networks are now central to many real-world applications including image classification~\cite{ResNet,VGGnet,xu2019positive}, object detection~\cite{ren2016object,SSD}, low-level vision~\cite{SRCNN}, and natural language processing (NLP)~\cite{hochreiter1997long,mikolov2010recurrent,vaswani2017attention}. Their performance usually depends on the availability of large amounts of finely labeled training data. For example, the ImageNet dataset~\cite{deng2009imagenet}, which is used for image classification, includes over 1.2M images in 1000 different categories; the MegaFace face recognition dataset~\cite{kemelmacher2016megaface} has over 1M photos from 690k unique users; and the CoQA dataset~\cite{reddy2019coqa} used for NLP contains over 127k questions with answers collected from over 8k conversations.

\begin{figure}[t]
	\centering
	\includegraphics[width=0.45\textwidth]{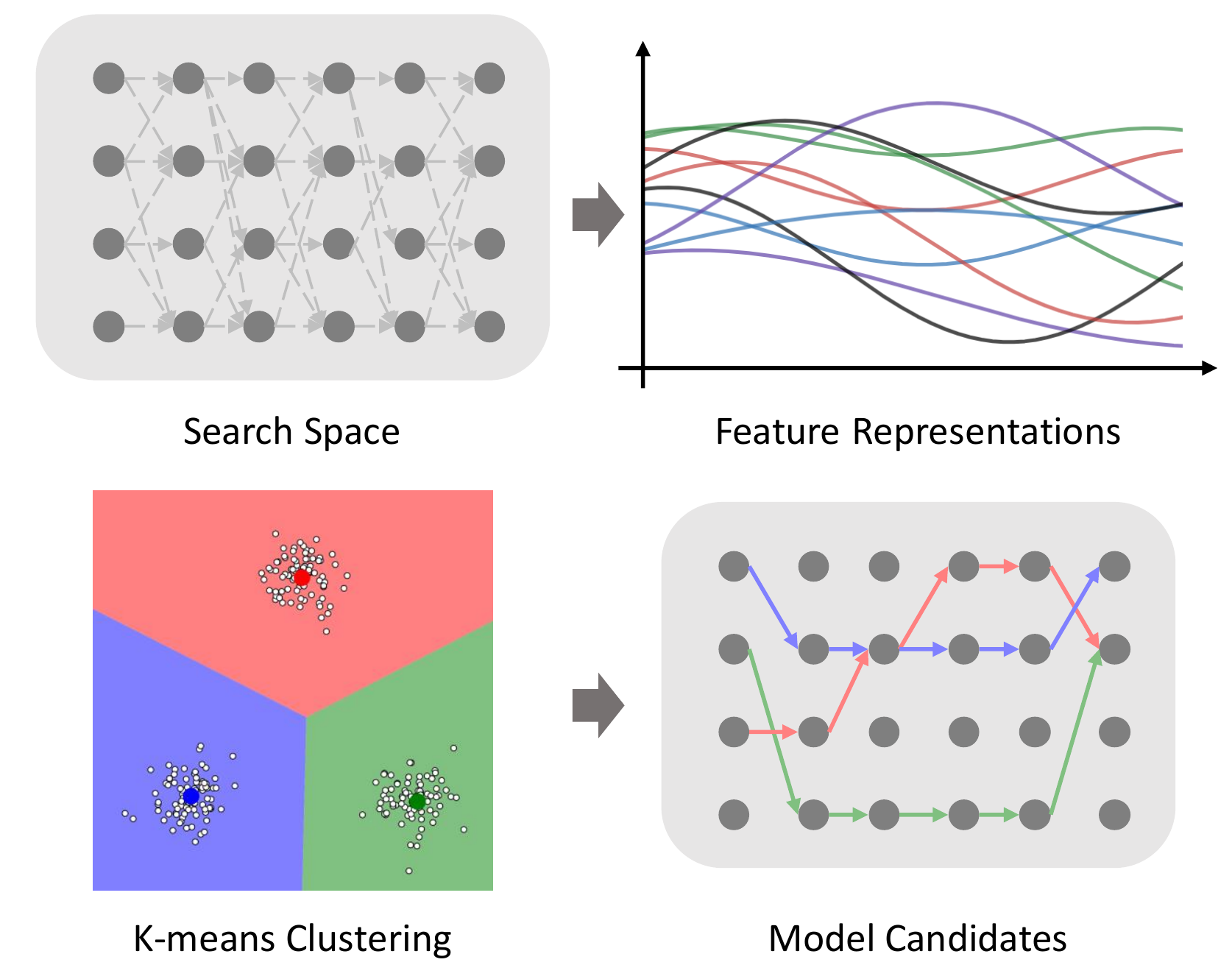}
	\caption{Diagram of the proposed DC-NAS. The given search space will be converted to a series of feature representations of different architectures. Then, a k-means algorithm is utilized to divide the searching problem, and the best architecture will be obtained by merging selected model candidates.}
	\label{Fig:pipeline}
\end{figure}

\begin{figure*}[t]
	\centering
	\begin{tabular}{ccc}
		\includegraphics[width=0.29\textwidth]{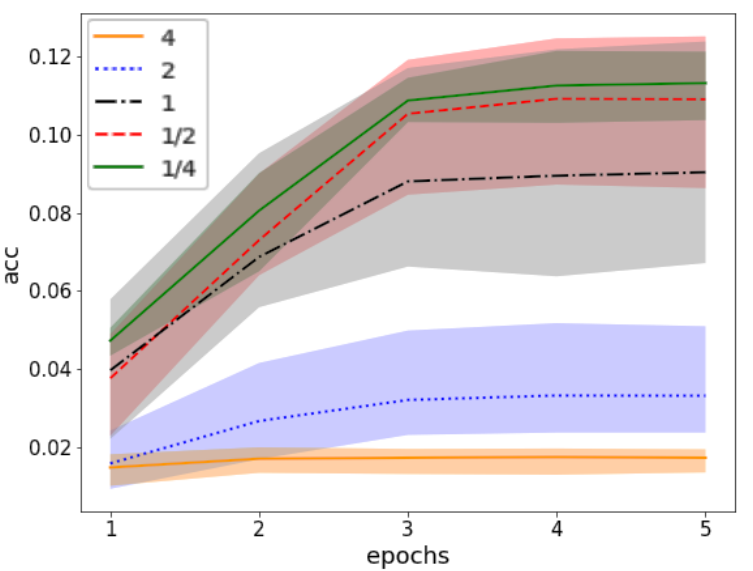}&
		\includegraphics[width=0.29\textwidth]{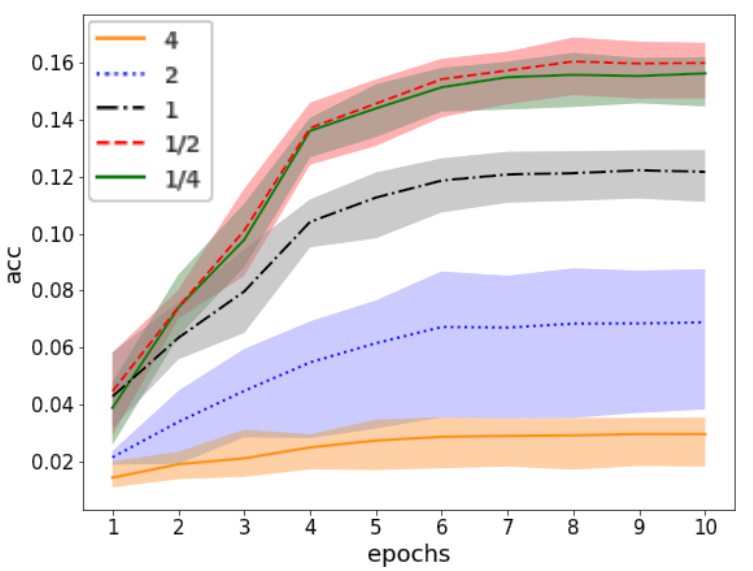}&
		\includegraphics[width=0.29\textwidth]{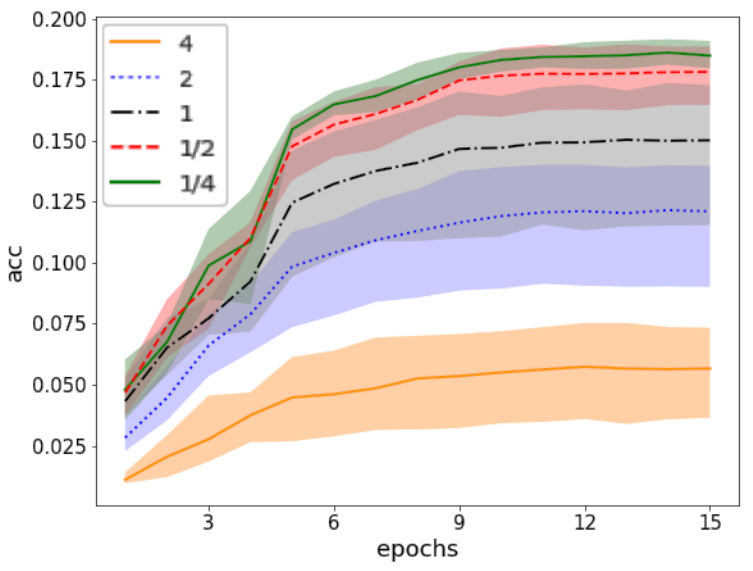}
		\\
		\small{(a)} & \small{(b)} & \small{(c)}\\
		\includegraphics[width=0.29\textwidth]{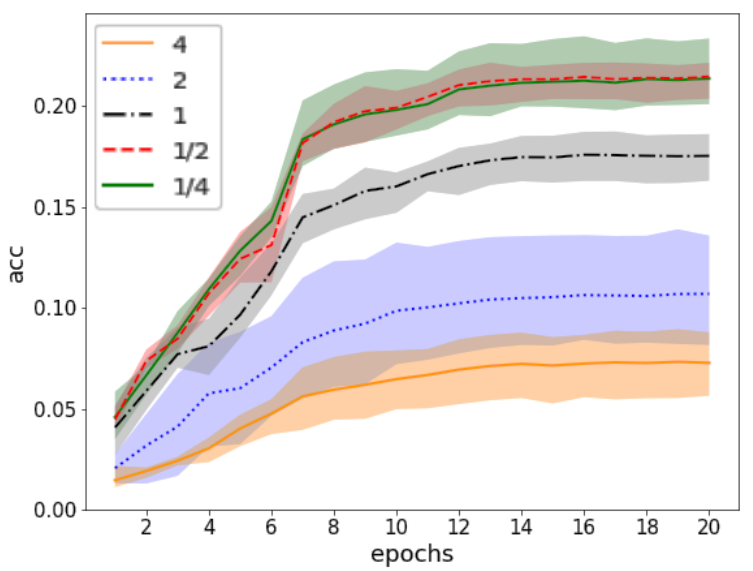}&
		\includegraphics[width=0.29\textwidth]{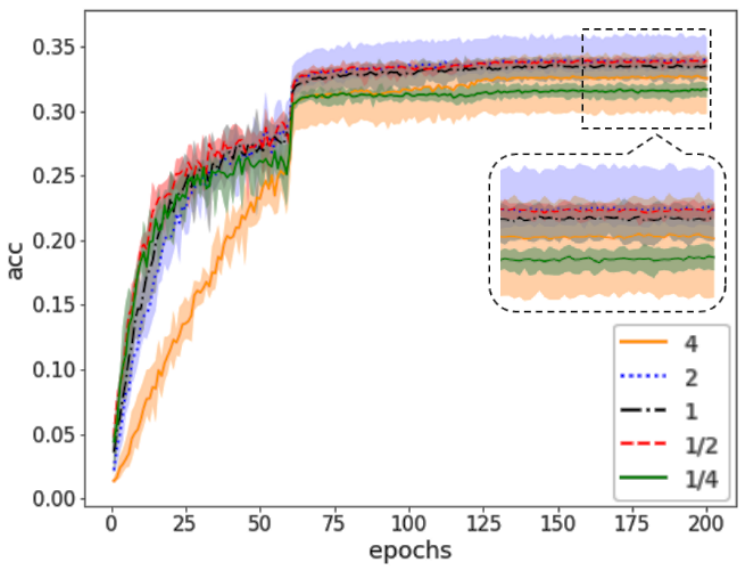}&
		\includegraphics[width=0.29\textwidth]{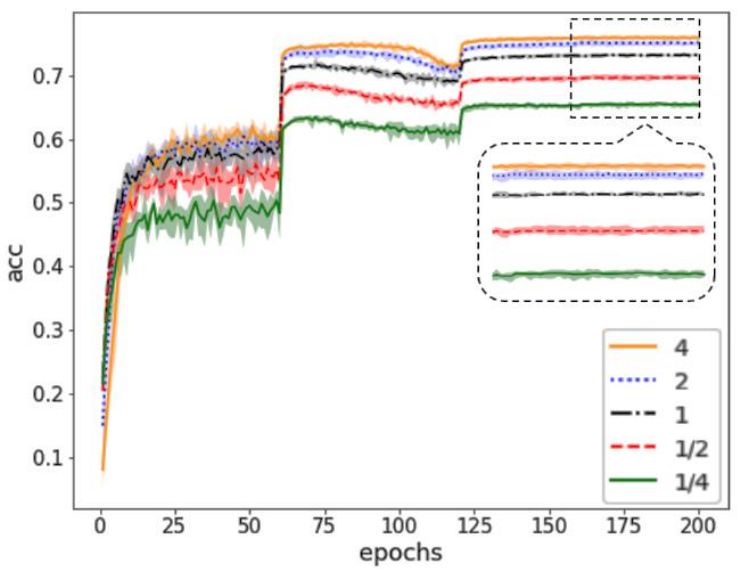}
		\\
		\small{(d)} & \small{(e)} & \small{(f)}\\
	\end{tabular}
	\caption{Illustrative results of searching different neural architectures using conventional early stopping. Five architectures with different widths (\ie, $4$, $2$, $1$, $\frac{1}{2}$, and $\frac{1}{4}$) are established in each panel. Their performance is tested on the CIFAR-100 validation set using different strategies. (a)-(e) are the results learned using different epochs on the $10\%$ training set, and (f) shows the exact performance of these networks.}
	\label{Fig:Intro}
\end{figure*}

In addition to large amount of data, well-designed architectures are critical for effective deep neural networks, especially convolutional neural networks (CNNs) for computer vision tasks. There are many network architectures. Simonyan and Zisserman presented VGGNet~\cite{VGGnet}, which contains over $10^9$ learnable parameters. He~\etal~\cite{ResNet} presented a shortcut operation for training deep neural networks with over $50$ layers. ResNeXt~\cite{ResNeXt} combines the shortcut operation with the 'split-transform-merge' operation by first splitting the channel of each layer, then applying transformation separately, before finally merging the outputs with an addition operation. Huang~\etal~presented DenseNet~\cite{huang2017densely}, which strengths information flow by adding a shortcut to each pair of blocks. In addition, some modern neural architectures have significantly lower computational costs, allowing their use on mobile devices. MobileNet~\cite{Mobilenet} reduces the computational complexity by replacing the traditional convolution operation with a combination of depth-wise convolution and point-wise convolution. ShuffleNet~\cite{Shufflenet} enables information flow between groups by adding a channel shuffle operation after group convolution. Mobilenet-v2~\cite{sandler2018mobilenetv2} further uses a linear bottleneck and an inverted residual block to reuse the features and to overcome the manifold collapse problem produced by the ReLU operation. Han~\etal~developed GhostNet~\cite{han2019ghostnet} and built efficient neural networks by generating more features from some inexpensive operations.

Despite the success of these networks and their variations, their design and implementation still require considerable human and computational resources. To address this, a number of neural architecture search (NAS) algorithms have been proposed to automatically search for optimal neural architectures in the given search space and dataset. Zoph \etal~\cite{zoph2016neural} generated the model descriptions of neural networks using a recurrent network and produced higher-accuracy architectures by training the recurrent neural network (RNN) with reinforcement learning. Real \etal~\cite{real2019regularized} used a modified evolutionary algorithm to search for better neural architectures in a given search space by introducing an age property to favor younger genotypes. Liu \etal~\cite{liu2018darts} represented the network architecture based on continuous relaxation, which allowed efficient architecture searching using gradient descent. Wang~\etal~\cite{wang2018towards} proposed an evolutionary method to automatically identify redundant filters in pre-trained deep neural networks. Pham \etal~\cite{pham2018efficient} reduced the search cost by searching for an optimal subgraph within a large computational graph with a controller.

While these NAS methods have significantly contribute to searching for better neural architectures, they have a significant limitation: in order to achieve an acceptable search cost, each searched architecture is not precisely optimized. To this end, early stopping is widely used in NAS for fast comparison. However, considering all sub-networks together in early stopping is highly unstable since they each have variable computational power. Figure~\ref{Fig:Intro} illustrates a toy experiment comparing five simple networks on the CIFAR-100 dataset with different early stopping settings. Since the convergence of models with fewer parameters and lower computational complexity is usually faster, the early stopping results in the entire search space are biased. In addition, a consensus was recently reached~\cite{sciuto2019evaluating,li2019random,yang2019evaluation} that random searching sometimes performs better than objective function-oriented NAS methods, also suggesting that the NAS evaluation procedure requires further refinement.

In this paper, we present a divide-and-conquer NAS (DC-NAS) scheme to better evaluate and handle the huge search space of neural architectures. In doing so, we highlight that the unified early stopping strategy for evaluating all networks sampled from the entire search space is unreliable, since the networks all have different numbers of parameters and computational power. We first observe the gradient change of each network's layers and operations and cluster them into several categories using the conventional k-means algorithm. Then, network comparisons are conducted in the same cluster by exploiting the usual early stopping strategy. Optimal networks in each cluster are finally merged and carefully trained to derive the best neural architecture in the given search space. The DC-NAS pipeline is shown in Figure~\ref{Fig:pipeline}. Since the gradient change can be directly calculated from the super-network with only limited cost and the k-means approach is also very efficient, DC-NAS does not obviously increase searching costs. Our extensive experimental results show that the proposed method can effectively solve the inaccurate evaluation problem and more efficiently perform NAS. Compared to state-of-the-art approaches, DC-NAS finds better performing networks in the same search space.

\section{Related Works}
To solve the problem of inaccurate evaluation of different networks, this work aims to divide the complex NAS problem into several sub-problems and obtain the optimal architecture by merging their results. Current NAS algorithms can be divided into two evaluation approaches: the early stopping strategy for evaluation and predictor based architecture searching.

\subsection{Early Stopping in NAS}
In order to automatically design high-performance deep neural networks, a series of algorithms have been proposed to search neural architectures in a given search space. Zoph~\etal~\cite{zoph2016neural} first proposed the concept of large-scale image classifier searching, which encoded different operations (\eg, convolution and pooling) and RNN connections to search for effective architectures using reinforcement learning (RL) by treating the accuracy of the current architecture as a reward and selecting policies to obtain the new architecture. Real~\etal~\cite{real2019regularized} proposed an evolutionary-based algorithm (EA) to search the architectures. Specifically, each individual neural network was regarded as an architectural component. Searching was performed across generations by using mutations and re-combinations of architectural components, with components showing better performance on the validation set picked and inherited by the next generation during evolution. Pham~\etal~\cite{pham2018efficient} developed efficient-NAS (ENAS), which significantly reduced the search cost by training a large computational graph and searching for an optimal subgraph that maximized the expected reward on a validation set using the weight-sharing method. 

Note that the reward used in RL and component performance in EA require training a large number of neural networks to convergence, which makes the search cost expensive. Thus, the early stopping strategy is commonly used to reduce the search cost, which trains the neural network over only a few epochs and uses the intermediate accuracy as the surrogate reward or performance of the component.

Early stopping is a very effective method for quickly evaluating or predicting the intermediate neural architectures during searches. However, effectiveness is compromised, since for different architectures, the relationship between the intermediate accuracy and the final accuracy is unclear. Specifically, taking different networks sampled from the entire search space and applying the same early stopping strategy is ineffective, since networks with more learnable parameters are usually harder to optimize.

In addition to the EA-based~\cite{yang2019cars} and RL-based NAS methods with early stopping, differentiable neural architecture searching (\eg, DARTS~\cite{liu2018darts}, Hit-Detector~\cite{guo2020hit}) also provides an efficient way to search deep neural networks. DARTS regarded the given super network as a continuous space and simultaneously maintained two sets of parameters with respect to the weights of the desired network and the weights for selecting different operations between nodes.
SNAS~\cite{xie2018snas} replaced the feedback mechanism triggered by constant rewards with gradient feedback from a generic loss. FBNet~\cite{wu2019fbnet} proposed a hardware-aware efficient conv-net designing method by integrating the network architecture latency into the loss function.

Differentiable NAS algorithms significantly increase the search speed by sharing weights across all sub-networks. However, determining the importance of each layer during the search is difficult. In other words, the parameter optimization results for the searched architectures are unreliable.

\subsection{Network Performance Predictor}
To avoid expensive search costs, a series of predictor-based methods have been proposed that better predict the latter part of the learning curve given the former part. For example, Domhan~\etal~\cite{domhan2015speeding} extrapolated network architecture performance from the first part of a learning curve by mimicking the early termination of bad runs using a probabilistic model. Klein~\etal~\cite{klein2016learning} studied the use of Bayesian neural networks to improve performance with a specialized learning curve layer.
Swersky~\etal~\cite{swersky2014freeze} assumed that the training loss during the fitting procedure roughly followed an exponential decay towards an unknown final value and proposed a regression model based on Bayesian optimization. Baker~\etal~\cite{baker2017accelerating} proposed a sequential regression model to predict the validation accuracy using features based on network architectures, hyper-parameters, and time-series validation performance data. 
However, good performance relies on the assumption that the learning curve should be smooth, which is often not the case in reality, since the learning curve may abruptly change when the learning rate changes. 

Some other predictors try to directly predict architecture performance before training~\cite{tang2020semi}. Deng~\etal~\cite{deng2017peephole} proposed using long short-term memory (LSTM) to predict the performance of a given architecture. Specifically, each layer of the architecture was embedded into a feature vector according to some predefined rules, and the vectors were then concatenated to produce the LSTM input. Sun~\etal~\cite{sun2019surrogate} encoded the network architecture by assuming that it is consisted of ResNet block and DenseNet block, extracted features based on it, and then predicted the accuracy with random forests. Xu~\etal~\cite{xu2019renasrelativistic} argued that the rankings between different models were much more important than absolute performance, and proposed a predictor based on a pairwise ranking-based loss function.

Although a well-trained neural architecture predictor can replace the early stopping approach and avoid the mixed evaluation problem, collecting the ground-truth labels (\ie, the exact network performance) to train the predictor is very expensive. Moreover, the predictor is sensitive to the number of training networks and the sampling strategy. Thus, NAS urgently requires an effective and accurate approach for comparing different searched architectures.

\section{Divide-and-conquer Searching}
We first revisit the evaluation of different sampled neural architectures in NAS, which is the main reason that some excellent architectures cannot be maintained during the search. Then, we develop divide-and-conquer NAS method. The original search space will be split to several clusters according to sub-network similarity.

\subsection{Evaluation Problem in NAS}
NAS represents a series of well-design approaches for obtaining the optimal neural architecture for the target task and related dataset from a huge search space. For a given search space $\mathcal S$ containing $p$ different neural architectures, \ie, $\mathcal{S} = \{\mathcal{N}_1,...,\mathcal{N}_p\}$, where $\mathcal N_i$ is the $i$-th sampled architecture from $\mathcal S$, each architecture is composed of a number of neural operations proven to be useful for deep learning (\eg, convolution, pooling, short-cuts, \etc.). Denoting the target dataset $\mathbf{D}$, the performance of each neural architecture can be calculated as
\begin{equation}
y_i = T(\mathcal{N}_i,\mathbf{D}),
\end{equation}
where $y_i$ is the accuracy of $\mathcal{N}_i$ on the test set and $T(\cdot)$ is the training function, which is tuned for the sampled network from $S$. In general, the training of a modern deep neural architecture is an intensive process (\eg, over eight GPU*days on ImageNet to train only one sub-network), and a large number of sub-networks exist in the search space. Evaluating all of these architectures individually is almost impossible. Thus, most NAS algorithms adopt an early stopping approach on a reduced dataset to evaluate each sub-network, \ie,
\begin{equation}
E(\mathcal{N}_i) = \tilde T(\mathcal{N}_i,\tilde{\mathbf{D}}),
\label{Fcn:early}
\end{equation} 
where $\tilde T(\cdot)$ is the training function with the early stopping strategy and $\tilde{\mathbf{D}}$ is the smaller dataset containing $\sigma$ (\eg, $5\%$) of the original training dataset to promote efficiency.

Although Eq.~\ref{Fcn:early} can significantly reduce the time taken to evaluate the searched networks, it is very hard to construct a smaller dataset and training method to satisfy the following objective function:
\begin{equation}
\min_E\frac{1}{p}\sum_{i=1}^p||E(\mathcal{N}_i)-y_i||_2^2,
\end{equation}
where $||\cdot||_2$ is the conventional $\ell_2$-norm. This is because deep neural networks are highly complex and hard to train on large-scale datasets, and there is no ranking-preserving relationship between early stopping performance and the final ground-truth accuracy of an arbitrary neural network. 

It is in fact unnecessary to predict the exact accuracy (or ground-truth accuracy) of the searched networks. If the order of all sub-networks can be preserved, the evaluation function $E(\cdot)$ can also accurately help us to identify networks with better performance, significantly improving NAS efficiency. Therefore, the objective function for optimizing the evaluating function can be written as
\begin{equation}
\min_E\sum_{i=1}^p\sum_{j=i+1}^p \text{sgn}(y_j-y_i)\times\text{sgn}(E(\mathcal{N}_i)-E(\mathcal{N}_j)),
\label{Fcn:obj}
\end{equation}
where $\text{sgn}(\cdot)$ is the sign function. Then, our goal is to find an efficient evaluator $E(\cdot)$ and embed it into mainstream NAS algorithms.

Figure~\ref{Fig:Intro} shows that applying the same early stopping strategy to all networks cannot achieve the desired functionality, because the optimization routes of the various networks are different. For instance, some light-weight neural networks will first achieve a relatively high accuracy but their subsequent improvements are not obvious. However, if some neural networks have very close representations, their convergence statuses are similar. Therefore, the global early stopping strategy is unsuitable for constructing $E(\cdot)$.

\subsection{Architecture Clustering}

As discussed in Eq.~\ref{Fcn:obj}, an ideal evaluation function is defined to compare different network architectures during the search. In fact, Eq.~\ref{Fcn:obj} can also be regarded as a predictive problem, with some existing works addressing this~\cite{deng2017peephole,sun2019surrogate,xu2019renasrelativistic}. However, these methods require a certain number of fully-trained networks sampled from the given search space to construct the performance predictor, which is also very resource intensive.

Handling the performance comparison task using a universal early stopping strategy is unreliable; nevertheless, early stopping is still the cheapest way to compare a set of architectures without too many differences. Therefore, we propose to conduct NAS in a divide-and-conquer manner to reduce the difficulty in comparing various architectures during the search. In practice, we divide all the sub-networks in the given search space $\mathcal{S}$ into $K$ categories using $K$ cluster centers to form $K$ different groups, \ie, $\mathbf{G}=\{\mathbf{G}_1,...,\mathbf{G}_K\}$. The corresponding objective function can be written as
\begin{equation}
\arg\min_{\mathbf{G}} \sum_{k=1}^K\sum_{\mathcal{N}\in\mathcal{G}_k} ||F(\mathcal{N})-F(\tilde{\mathcal{N}}_k)||_2^2,
\label{Fcn:arc_cl}
\end{equation}
where $\tilde{\mathcal{N}}_k$ denotes the $k$-th clustering center of group $\mathbf{G}_k$, and $F(\cdot)$ is the feature representation of each input deep neural architecture. 

Establishing the relationship between deep neural architectures and their performance is difficult, and how to represent deep networks is also extremely complicated due to the various structures and billions of trainable parameters. Fortunately, the current mainstream search space is often regularized to a super-network~\cite{liu2018darts,xie2018snas,wu2019fbnet}, \ie, the depths and types of connections are usually fixed, and all the operations needing to be searched are merged in it. Thus, we suggest representing each sampled network from the given search space $\mathcal{S}$ using a code with fixed length $L$, \ie, the number of layers.

One straightforward way to represent neural architectures is to regard the operations in each layer (\eg, ``conv'' and ``ReLU''~\cite{deng2017peephole,sun2019surrogate}) or computational complexity (\ie, FLOPS and memory~\cite{xu2019renasrelativistic,istrate2019tapas}) as neural architecture features. However, directly using these features does not capture the relationship between any two layers and operations, and the computational complexity cannot be computed for some parameter-free layers, \eg, pooling and short-cut layers. In addition, the phenomenon shown in Figure~\ref{Fig:Intro} motivates us to encode the optimization difficulty of different neural architectures. Therefore, given $L$ as the number of layers in the neural architectures and $\eta$ as the maximum epochs trained on the training set, we propose using the following representation function to encode an arbitrary network $\mathcal{N}_i$ into an $(L\times\eta)$-dimensional matrix, \ie,
\begin{equation}
\hat F_{l,e}(\mathcal{N}_{i})=\frac{<{\theta_{i}^{l,1}}, {\theta}_{i}^{l,e}>}{||\theta_{i}^{l,1}||\cdot||{\theta}_{i}^{l,e}||},
\label{Fcn:pfeat}
\end{equation}
where $\theta_{i}^{l,e}$ are the parameters in the $l$-th layer in network $N_i$ trained on the reduced training set $\tilde{\mathbf{D}}$ after $e$ epochs. The above function calculates the changes in parameters of a given network between the first and the $e$-th epochs, which can reflect the sensitivity of each layer in the sampled network. For instance, a layer with fewer parameters will converge faster than another layer with more parameters due to the different optimization difficulty, as shown in Figure~\ref{Fig:Intro}. Thus, we can use these features to distinguish neural architectures with different properties and divide the huge search space $\mathcal{S}$ accordingly.

However, Eq.~\ref{Fcn:pfeat} also faces the same problem of not being able to represent some layers (such as pooling and short-cut) without trainable parameters. Thus, we further extend it to the output of each layer as the feature representation of a given neural network, \ie,
\begin{equation}
F_{l,e}(\mathcal{N}_{i})=\frac{<{o_{i}^{l,1}}, {o}_{i}^{l,e}>}{||o_{i}^{l,1}||\cdot||{o}_{i}^{l,e}||},
\label{Fcn:feat}
\end{equation}
where $o_{i}^{l,e}$ is the averaged output feature of the $l$-th layer in network $\mathcal{N}_{i}$ for a certain amount of data trained on $\tilde{\mathbf{D}}$ with $e$ epochs. To save the calculation of Eq.~\ref{Fcn:feat} and balance the information of each dimensionality, we further apply a global average pooling on the feature representation ${o}$ of each layer before calculating $F$.

Then, we can utilize conventional k-means clustering or hierarchical clustering ~\cite{yaari1997segmentation,alsabti1997efficient} to divide the original search space into $K$ clusters according to the feature similarity between any two sampled neural architectures.

\begin{algorithm}[t]
	\renewcommand\baselinestretch{1.1}\selectfont
	\caption{DC-NAS for Searching Neural Architectures.}
	\label{Alg:main}
	\begin{algorithmic}[1]
		\Require An arbitrary search space $\mathcal{S}$ containing $p$ sub-networks $\{\mathcal{N}_1,...,\mathcal{N}_p\}$ with $L$ layers. The target dataset $\mathbf{D}$ and it's shrunk version $\tilde{\mathbf{D}}$, the clustering number $K$, the number of epochs $\eta$ for extracting feature representations, and the number of sampled networks $s\leq p$.
		\State \textbf{Search Space  Clustering:}
		\State Randomly generate $s$ neural architectures from $\mathcal{S}$;
		\For{$i = 1$ to $s$}
		\State Initialized the parameters in $\mathcal{N}_i$;
		\State Train the network $\mathcal{N}_i$ on $\tilde{\mathbf{D}}$ with $\eta$ epochs;
		\For{$l = 1$ to $L$}
		\For{$e = 1$ to $\eta$}
		\State Obtain features $o_{i}^{l,e}$ of the $l$-th layer after training $e$ epochs;
		\State Calculate $F_{l,e}(\mathcal{N}_{i})$ using Eq.~\ref{Fcn:feat};
		\EndFor
		\EndFor
		\EndFor
		\State Cluster $s$ networks into $K$ groups $\{\mathcal{G}_1,...,\mathcal{G}_K\}$ using Eq.~\ref{Fcn:arc_cl};
		\State \textbf{Merging Architectures:}
		\For{$i = 1$ to $K$}
		\State Find the best architecture $\hat{\mathcal{N}}_i$ in $\mathcal{G}_i$ using the conventional early stopping (Eq.~\ref{Fcn:res1} for architectures trained separately, Eq.~\ref{Fcn:res3} for architectures sampled from a super network);
		\EndFor
		\Ensure The optimal network $\mathcal{N}^*$ among the $K$ groups.
	\end{algorithmic}
\end{algorithm}

\subsection{Divide-and-conquer Network Comparison}
After obtaining $K$ deep neural architecture clusters, the evaluation difficulty is naturally reduced, since each cluster represents deep models with similar optimization paths and performance. Therefore, we propose to first distinguish the best architecture in each group and then merge them to produce the final result.

For architectures in the same cluster, we can reuse the conventional early stopping strategy to distinguish searched architectures in the same cluster with similar gradient changes. Therefore, for $K$ clusters, the $K$ optimal architecture can easily be solved by using
\begin{equation}
\hat{\mathcal{N}}_k = \mathop{\arg\max}_{\mathcal{N}\in\mathcal{G}_k}E(\mathcal{N}) = \mathop{\arg\max}_{\mathcal{N}\in\mathcal{G}_k}\tilde T(\mathcal{N},\tilde{\mathbf{D}}),
\label{Fcn:res1}
\end{equation}
where $\hat{\mathcal{N}}_k$ is the best network architecture in the $k$-th group selected by the reduced dataset $\tilde{\mathbf{D}}$.

Eq.\ref{Fcn:res1} is suitable when evaluating different neural networks that have been trained separately. However, for the sub-networks extracted from the pre-trained super network, the performance of each sub-network without further training is usually poor and might not represent its true fitting ability. Note that in a given pre-trained super-network, the operation parameters in each layer give the probabilities of choosing different operations in that layer. Thus, we use the sub-network with the largest probability of being chosen in a cluster as the representation of this cluster. Given ${ a(\mathcal N)}=\{a(\mathcal N^1),a(\mathcal N^2),\cdot\cdot\cdot,a(\mathcal N^L)\}$ as the operation probability parameters of each layer in a sub-network $\mathcal N$, we have:
\begin{equation}
\hat{\mathcal{N}}_k = \mathop{\arg\max}_{\mathcal{N}\in\mathcal{G}_k}\sum_{l=1}^L{a(\mathcal N^l)},
\label{Fcn:res3}
\end{equation}

Then, the optimal neural architecture in the given search space $\mathcal{S}$ can be easily derived using the following function:
\begin{equation}
\mathcal{N}^* = \mathop{\arg\max}_{\hat{\mathcal{N}}_i}T(\hat{\mathcal{N}}_i,\mathbf{D}), \quad \forall\;\; i = 1,...K,
\label{Fcn:res2}
\end{equation}
where, the performance of each network selected in the specific cluster will be derived by fully training them on entire dataset $\mathbf{D}$ for an accurate comparison. 

Note that although the training procedure on $\mathbf{D}$ is somewhat expensive, the number of clusters $K$ is relatively small and Eq.~\ref{Fcn:res2} does not dominate our scheme. The detailed searching procedure of DC-NAS for accurately searching neural architectures is summarized in Algorithm~\ref{Alg:main}.

\section{Experiments}
We have therefore developed a novel NAS framework using the divide-and-conquer approach, splitting the huge search space into several groups to reduce the difficulty in evaluating the searched neural architectures. Here we verify the effectiveness of the proposed method on several NAS search spaces.

\subsection{Validations on Toy Search Space}

\paragraph{\textbf{Search Space Definition.}} We first validate the proposed method using a very simple search space, as shown in Figure~\ref{Fig:Intro}.
The baseline network includes six convolutional layers and a fully-connected layer. We extend this baseline network by adjusting the channel ratio in different layers, \ie, $\frac{1}{4}$, $\frac{1}{2}$, $1$, $2$, $4$. Thus, the entire search space contains $5^6 = 15625$ neural architectures with different parameters and computational complexity. CIFAR-100~\cite{krizhevsky2009learning} is selected as the target dataset.

Since the search space is relatively small, we first train all of these networks on $10\%$ of CIFAR-100, \ie, $\tilde{\mathbf{D}}$ in Eq.~\ref{Fcn:early}. The number of epochs used in the original baseline network is 200. For efficiency, these architectures are trained separately for 20 epochs. Then, all the images in the smaller training dataset $\bf\tilde D$ are used to extract the feature representations of each layer of these architectures after training different epochs. 

\paragraph{\textbf{Architecture Clustering and Searching.}} 
After obtaining the features of these neural architectures, we obtain a feature tensor $\mathcal T\in \mathbb R^{s\times L\times\eta}$ in which $L=6$. Specifically, we set $s=15625$ and $\eta\in\{5,10,20,30\}$. Then, we apply the k-means algorithm on this matrix to generate a series of architecture groups. After that, we evaluate each network cluster using Eq.\ref{Fcn:res1}, and $K$ selected networks will be fully trained on the entire training dataset $\bf D$ to obtain the final result, \ie, the optimal architecture in the given search space.

\renewcommand{\multirowsetup}{\centering}
\begin{table*}[t]
	\begin{center}
		\renewcommand\arraystretch{1.0}
		\caption{Searched results of DC-NAS with different clustering number $K$ and evaluating epochs $\eta$ on the toy search space.}
		\begin{tabular}{l|c|c|c|c|c|c}
			\hline
			DC-NAS & $K = 1$ & $K = 3$ & $K = 5$ & $K = 10$ & $K = 20$ &$K = 50$ \\
			\hline\hline
			Top-1 acc(\%) (RS) & 75.01 & 75.36 & 75.49 & 75.58 & 75.71 & 75.83\\
			\hline
			Top-1 acc(\%) ($\eta = 5$)  & 71.74 & 73.96 & 73.96& 73.96& 75.38& 75.47 \\
			\hline
			Top-1 acc(\%) ($\eta = 10$) & 71.13 & 75.42 & 75.42& 75.42& 75.83& 75.83\\
			\hline
			Top-1 acc(\%) ($\eta = 20$) & 73.32 & 75.91 & 75.91 &75.91 & 75.91&76.07\\
			\hline
			Top-1 acc(\%) ($\eta = 30$) & 74.07 & 75.91 & 75.91 &75.91 & 75.91&76.07\\
			\hline
			
		\end{tabular}
		\label{Tab:kmeans}
	\end{center}
\end{table*}

In Table~\ref{Tab:kmeans}, we test the impact of parameter $K\in\{1,3,5,10,20,50\}$ in dividing the search space according to the feature representations of the architectures. We also examine the impact of parameter $\eta$ ranging from $\{5,10,20,30\}$. When $K=1$, DC-NAS is equivalent to the conventional early stopping strategy for evaluating searched neural architectures, which is used in almost all NAS algorithms. An increasing $K$ is more likely to obtain the global optimum of the search space, while increasing the evaluation time. When $K=15625$, the proposed method degrades to a traversal search in which the global optimum is guaranteed, but the evaluation time is unaffordable. We further compare the effectiveness of the proposed method with random search (RS). Note that we have trained $s=15625$ different architectures on $10\%$ of the original dataset with $10\%$ of the original epochs. Thus, for a given parameter $K$, the RS method randomly selects $156+K$ different neural architectures from the search space, and the architectures are fully trained to obtain the validation result. The architecture with the best validation result is selected, and this experiment is repeated $20$ times to overcome randomness.

As shown in Table~\ref{Tab:kmeans}, the best search results are achieved when $\eta=30$, which is slightly higher than that of $\eta=20$ only when $K=1$. On the one hand, a larger $\eta$ indicates a higher dimension of the feature representation which increases the amount of information to help generate better clusters. On the other, a larger $\eta$ alleviates the randomness of the feature representation, and the feature can more precisely represent the degree of convergence of the layer. However, a larger $\eta$ will also increase the search cost of the proposed DC-NAS algorithm, and the architecture search problem is exactly a method of exhaustion when $\eta \rightarrow \inf$. Thus, we suggest to set $\eta = 20$ (\ie, $10\%$ of the epochs for fully training) for a trade-off between the performance of neural architecture and search cost.

\begin{figure*}[t]
	\centering
	\subfloat[K=1]{
		\begin{minipage}[t]{0.28\linewidth}
			\centering
			\includegraphics[width=0.63\linewidth]{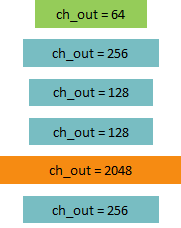}
		\end{minipage}
	}
	\centering
	\subfloat[K=3]{
		\begin{minipage}[t]{0.56\linewidth}
			\centering
			\includegraphics[width=0.99\linewidth]{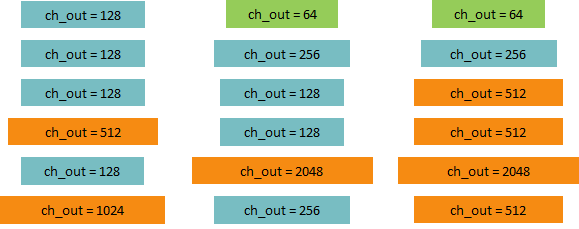}
		\end{minipage}
	}
	
	\centering
	\subfloat[K=5]{
		\begin{minipage}[t]{0.84\linewidth}
			\centering
			\includegraphics[width=1\linewidth]{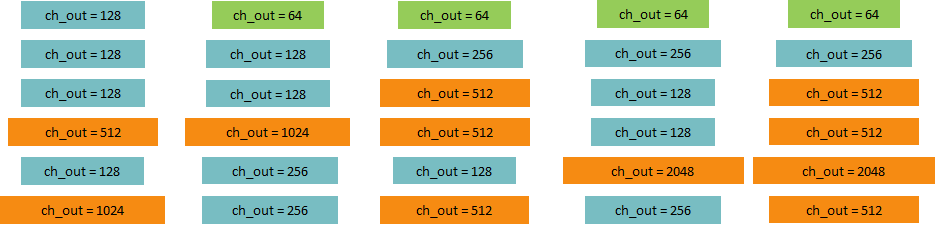}
		\end{minipage}
	}
	\caption{Visualization of the best deep neural architectures in each cluster with $K=1$, $K=3$ and $K=5$, respectively.}
	\label{figure_4}
\end{figure*}

Moreover, $K$ is also a very important parameter in our method. When $K=1$, the random search method achieves the best result, which means that the conventional early stopping method is not comparable to RS. However, the proposed method delivers significant gains even with a small cluster number ($K=3$), improving the search result from $73.32\%$ to $75.91\%$ and outperforming the RS method. This shows that DC-NAS can achieve a promising local optimum with only a slightly increase in the evaluation cost. We found that all the selected networks $\{\hat{\mathcal N}_k\}_{k=1}^K$ with a smaller $K$ also appear when $K$ increases as shown in Figure~\ref{figure_4}. The best network searched using $K=5$ includes the results using $K=3$, which demonstrates that the performance of the searched architecture will not decrease when $K$ becomes larger. In addition, we also find that a larger $K$ will not obviously enhance the accuracy of the searched network, since the number good candidates with sufficient discrimination \wrt the number of intrinsic clusters is limited, and we keep $K=5$ in the following experiments.

In order to explicitly understand the architecture clustering procedure, we detail the selected networks in each cluster using intra-cluster early stopping with $K = 5$ and $\eta=20$ (Table~\ref{Tab:resp}). Meanwhile, the searched architectures using different $K$ are also drawn in Figure~\ref{Fig:Intro}, selected networks in different clusters present variant architectures. Revising Figure~\ref{Fig:Intro}, for the given search space, the conventional early stopping method will search for the architecture with a small channel ratio, \ie, network architectures with few parameters and FLOPs, since a smaller network will converge faster on reduced dataset $\bf\tilde D$ at the beginning of training, thus outperforming larger networks with better ground-truth performance when applying the early stopping strategy. However, in Table~\ref{Tab:resp}, we show that there is an approximately $6$-times difference in the parameters and FLOPs between the largest and smallest models selected by DC-NAS, demonstrating that the proposed method can effectively solve the problem shown in Figure~\ref{Fig:Intro}. Furthermore, it can be found in Table~\ref{Tab:resp}, $\tilde{N}_3$ is about $1.87\times$ larger than $\tilde{N}_2$ with a slightly lower performance. This result illustrates that a larger model does not necessarily have higher accuracy. Overall, the proposed DC-NAS can distinguish models with different optimization difficulties and accurately evaluate them to obtain better architectures.

\begin{table}[t]
	\begin{center}
		\renewcommand\arraystretch{1.2}
		\caption{Optimal architectures in each cluster when $K = 5$.}
		\begin{tabular}{c|c|c|c|c}
			\hline
			Network & Channel Ratio & Param & FLOPs & acc(\%) \\
			\hline\hline
			$\tilde{N}_2$ & $4,2,1,2,0.25,2$ & 4.30M & 0.46G  & 73.64\\
			\hline
			$\tilde{N}_1$ & $2,2,1,4,0.5,0.5$ & 4.76M & 0.61G & 73.50\\
			\hline
			$\tilde{N}_4$ & $2,4,4,2,0.25,1$ & 5.69M & 1.14G & 73.60\\
			\hline
			$\tilde{N}_3$ & $2,4,1,0.5,4,0.5$ & 8.08M & 0.72G & 73.32\\
			\hline
			$\tilde{N}_5$ & $2,4,4,2,4,1$ & 23.38M & 2.27G & 75.91 \\
			\hline
		\end{tabular}
		\label{Tab:resp}
		\vspace{-1.5em}
	\end{center}
\end{table}

In addition, we sometimes need to sample some examples from the smaller dataset $\bf \tilde D$ in order to reduce the time taken to compute the averaged output feature of a specific layer in a given network (Eq.~\ref{Fcn:feat}). In order to verify the sensitivity of the proposed method in choosing different examples for extracting features, we repeat the search $10$ times with $K = 5$ and $\eta = 20$, each time randomly sampling 1000 images from the reduced training dataset $\bf \tilde D$ accordingly. The largest gap between the accuracy of the best and worst searched network is only about $0.2\%$, showing that the feature representation generation method utilized in Eq.~\ref{Fcn:feat} is very robust for distinguishing sub-networks in the given search space.

\begin{table*}[t]
	\begin{center}
		\small
		\renewcommand\arraystretch{1.0}
		\caption{Searched results of DC-NAS and state-of-the-art methods on ImageNet.}
		\vspace{-0.5em}
		\begin{tabular}{c|cc|ccc|cc}
			\hline
			\multirow{2}{*}{Method} & Search & Search & Latency & FLOPs & Params&Top-1 & Top-5 \\
			& methods & space & (ms) & (M) & (M) & acc(\%) & acc(\%)\\
			\hline\hline
			Random Search &-&-&-&276&4.3&72.0&90.6\\
			MobileNetV1~\cite{howard2017mobilenets} &handcraft & -  & -&- & 4.2 & 70.6 & 89.5 \\
			MobileNetV2 1.0$\times$~\cite{sandler2018mobilenetv2} & handcraft &- &-& 300 & 3.4 & 72.0 & 91.0\\
			ShuffleNetV1 1.5$\times$ (g=3)~\cite{zhang2018shufflenet} & handcraft &  &-& 292 & -    & 71.5 & -\\
			ShuffleNetV2 1.5$\times$~\cite{ma2018shufflenet} & handcraft & -&-& 299 & 3.5 & 72.6 & - \\
			MobileNetV2 1.4$\times$~\cite{sandler2018mobilenetv2} & handcraft &-& -& 585 & 6.9 & 74.7 & - \\
			ShuffleNetV2 2.0$\times$~\cite{ma2018shufflenet} & handcraft & -&- & 591 & 7.4 & 74.9 & - \\
			\hline
			ChamNet-B~\cite{dai2019chamnet} & predictor & layer-wise  &-& 323 & - & 73.8 & -\\
			NASNet-A~\cite{zoph2018learning} & RL & cell &-& 564 & 5.3 & 74.0 & - \\
			MnasNet~\cite{tan2019mnasnet} & RL & stage-wise & -&317 & 4.5 & 74.0 & -\\
			FBNet-B~\cite{wu2019fbnet} & gradient & layer-wise &  87.07&295 & 4.5 & 74.1 & - \\
			FBNet-B (our impl.) & gradient & layer-wise & 104.25& 326 & 4.7 & 73.7 & 91.5 \\
			ProxylessNAS-R~\cite{cai2018proxylessnas} & gradient & layer-wise &-& -& 5.8 & 74.6 & 92.2\\
			MnasNet-92~\cite{tan2019mnasnet} & RL & stage-wise& -&388 & 4.4 & 74.8 & -\\
			FBNet-C~\cite{wu2019fbnet} & gradient & layer-wise &102.83& 375 & 5.5 & 74.9 & - \\
			FBNet-C (our impl.) & gradient & layer-wise &116.55& 406 & 5.5 & 74.8 & 92.1 \\
			\hline			
			DC-NAS-A &  & \multirow{5}{*}{layer-wise} & 117.89 &319 & 4.7 & 73.4& 91.5\\
			DC-NAS-B & gradient &  & 84.19 &328 & 4.9 & 73.7& 91.5\\
			DC-NAS-C & + & & 104.10&341 & 5.0 & 74.2& 91.6\\
			DC-NAS-D & cluster & & 103.39&369 & 4.9 & 74.7& 92.1\\
			DC-NAS-E &  & & 118.12&428 & 5.5 & \textbf{75.1}& \textbf{92.2}\\
			\hline
		\end{tabular}
		\label{Tab:imgnet}
	\end{center}
\end{table*}

\subsection{Validations on ImageNet}
After conducting experiments on CIFAR-100 data and analyzing the impact of each parameters in the proposed DC-NAS, we further employ the new method on the challenging large-scale ImageNet dataset~\cite{ImageNet}. This dataset contains over 1.2M images from 1000 categories. FBNet~\cite{wu2019fbnet} is selected as the baseline method because of the excellent performance on both model accuracy and latency.

\paragraph{\textbf{Search Space Definition.}} We use the same search space proposed in FBNet~\cite{wu2019fbnet}, which is a layer-wise search space with a fixed super-network. The super-network architecture is composed of a $3\times3$ convolution layer, followed by seven SB blocks, a $1\times1$ convolution layer, a $7\times7$ average pooling layer and a fully-connected layer. Here SB block is the block that needs to be searched. Specifically, the block structure is fixed as a $1\times1$ convolution followed by a $k\times k$ depthwise convolution and another $1\times1$ convolution. ReLU activation is used after each layer, except for the last $1\times1$ convolution. When the stride of the block $s=2$, the stride of the first $1\times1$ convolution is 2. When $s=1$, a skip connection is added between the input and the output of the block.

When searching for the SB block, the expansion ratio $e$, which determines the channel size expansion from input to output of the first $1\times1$ convolution, can be searched from $e\in\{1,3,6\}$, and the kernel size $k$ of the depthwise convolution is $k\in\{3,5\}$. Furthermore, the group convolution followed by a channel shuffle operation can be used in the first and last $1\times1$ convolutions. Finally, a skip operation can be applied to the SB block, which actually cancels the block and reduces the architecture depth. There are $9$ different candidate operations forming a search space of size $9^{22}$.

\paragraph{\textbf{Results on ImageNet.}} Following~\cite{wu2019fbnet}, we first train the super-network on ImageNet-100 with 90 epochs. The first 10 epochs are trained with fixed operation parameters and the next 80 epochs are trained normally. In each epoch, the operator weight $w$ is first trained on $80\%$ of ImageNet-100 using SGD, and the probability parameters $a$ are trained on the remaining $20\%$ of the training set using Adam with an initial learning rate of 0.01 and weight decay of 0.0005. We then sample $10^7$ different architectures from the search space according to the operation probability of the pre-trained super-network. Specifically, we normalize the probabilities of the operations in the same layer, and then select operation of that layer based on the normalized probability. After that, we extract features using Eq.~\ref{Fcn:feat} with $10000$ images sampled from the ImageNet-100 training set. Finally, the feature representations of $10^7$ different samples are clustered into $K$ different groups using the k-means algorithm, and the representative sub-network from each group is selected using Eq.~\ref{Fcn:res3}. Considering the evaluation cost, we manually set a small cluster number $K=5$ in our experiment. When increasing $K$ to a larger number (\eg, $K=10$), we do not discover a significant increase of the search result. The five representative sub-networks are fully trained on ImageNet, and the search results are shown in Table~\ref{Tab:imgnet}, \ie, DC-NAS-A to DC-NAS-E. 

\paragraph{\textbf{Compare to state-of-the-art methods.}}  The results of the proposed DC-NAS are compared to state-of-the-art models designed manually and automatically. The evaluation metrics are top-1/top-5 accuracies on the ImageNet validation set, and FLOPs and parameters of the searched models. During the experiment, we found that the results of FBNet~\cite{wu2019fbnet} reported in the original paper is different from the results that is implemented by ourselves, which was also pointed out in ~\cite{stamoulis2019single}. Thus, we report both the evaluation metrics of the searched results mentioned in the original paper and the results implemented by us (denote as `our impl.'). Table~\ref{Tab:imgnet} shows that our DC-NAS achieves a top-1 accuracy of \textbf{$75.1\%$}, which is a new stat-of-the-art ImageNet accuracy among hardware-efficient NAS models. Meanwhile, we also compare the proposed DC-NAS with a number of state-of-the-art methods including NASNet~\cite{zoph2018learning}, MnasNet~\cite{tan2019mnasnet}, ProxylessNAS~\cite{cai2018proxylessnas}, \etc. The comparison results shown in Table~\ref{Tab:imgnet} also demonstrate that the proposed DC-NAS outperforms other methods under the same range of FLOPs.

\paragraph{\textbf{Search cost.}} The training process of super-net is exactly the same as in FBNet~\cite{wu2019fbnet}, which is 216 GPU hours. In this experiment, we do not need to adjust the hyper-parameter $\eta$. Extract features of $10^7$ different architectures using Eq.~\ref{Fcn:res3} takes about 15 GPU hours. Finally, clustering architectures into different groups by applying k-means on the extracted features can be done in a few minutes. Thus, the total search cost of DC-NAS is 231 GPU hours, which is $1.07\times$ compared to that of the baseline FBNet~\cite{wu2019fbnet}. In addition, it is easy to embed the proposed divide-and-conquer strategy into other frameworks to obtain better performance.

\paragraph{\textbf{Latency.}} Besides the FLOPS and model sizes, we also report the latency of different model on an ARM based mobile device for a fair comparison. The latency of the DC-NAS-E with the highest performance for predicting a $224\times 224$ image is $118.12ms$, which is very close to that ($116.55ms$) of the FBNet-C. At the same time, we can find from Table~\ref{Tab:imgnet}, the latency of a neural network is not linearly related to FLOPS and model size. Thus, the hardware-aware strategy is essential for neural architecture search. Overall, the proposed DC-NAS can provide networks with the state-of-the-art performance on the given search space.

\section{Conclusions}
In this paper we present a divide-and-conquer neural architecture search (DC-NAS) approach for effectively searching deep neural architectures. The proposed method overcomes the evaluation problem inherent in traditional NAS methods, namely they choose architectures that perform well on smaller datasets using the early stopping strategy while rejecting architectures that are actually acceptable after full training. Specifically, DC-NAS first extracts features that represent the speed of convergence of sub-networks according to the change in output features of each layer after each training epoch. It then calculates the similarity between two different architectures according to the feature representations. After that, traditional k-means clustering is used to divide the huge search space into several groups, and the best architectures in each group are further compared to obtain the best searched architecture. Our experimental results show that the proposed DC-NAS can significantly improve the performance of the searched architecture with only a slightly increase in the evaluation cost compared to traditional NAS methods.

\bibliography{ref}
\bibliographystyle{ieee}

\end{document}